\documentclass{llncs} 

\usepackage{multicol}
\usepackage[latin1]{inputenc}
\usepackage[english]{babel}
\usepackage{epic}
\usepackage{epstopdf}
\usepackage{url}
\usepackage{mathptmx}
\usepackage{indentfirst}
\usepackage{amssymb}
\usepackage{amsmath}
\usepackage{textcomp}
\usepackage{wrapfig}
\usepackage{natbib}
\usepackage{proof}
\usepackage[pdftex]{graphicx}

\usepackage{prooftree} 
\usepackage{proof}

\usepackage{array}
\newcommand{\editout}[1]{}

\newcommand\implique\Rightarrow
\newcommand\imp\rightarrow 
\newcommand\flsshort\rightarrow 
\newcommand\fls\rightarrow 
\newcommand\lfrom{\mathop{/}}
\newcommand\lto{\mathop{\backslash}} 
\newcommand\existe\exists
\newcommand\qqst\forall 
\newcommand\ou\lor 

\renewcommand\l\lambda 

\input{qobitree}

\newcommand\type[1]{^{#1}}
\newcommand\et\land
\newcommand\fl{\rightarrow}

\newcommand\ma[1]{``\emph{#1}''}

\newcommand\ttt{t}
\newcommand\eee{e}

\usepackage{gb4e} 

\begin{document}

\title{
Natural {L}anguage {S}emantics and {C}omputability 
}
\author{Richard Moot \inst{1} \and Christian Retor\'e\inst{2}}
\institute{CNRS LaBRI \and Universit\'e de Montpellier \& LIRMM}
\authorrunning{R. Moot, Ch. Retor\'e}
\maketitle

\begin{abstract}
This paper is a reflexion on the computability of natural language semantics. It does not contain a new model or new results in the formal semantics of natural language: it is  rather a computational analysis of the logical models and algorithms currently used in natural language semantics,
defined as the mapping of a statement to logical formulas ---
formulas, because a statement can be ambiguous. We argue that as long
as possible world semantics is left out, one can compute the semantic
representation(s) of a given statement, including aspects of lexical
meaning. We also discuss the algorithmic complexity of this
process. 
\end{abstract}
 
\section*{Introduction}

In the well-known Turing test for artificial intelligence, 
a human
interrogator needs to decide, via a question
answering session with two terminals, which of his two interlocutors
is a man and which is a machine \citep{turing50}. 
Although early systems like Eliza based on matching
word patterns may seem clever at first sight,
\editout{\footnote{The modern
  question answering systems found on many mobile telephones today
  essentially push the Eliza methodology far enough to be useful in
  many circumstances, although the smart interrogator can expose
  the limitations of these systems, by getting the system to respond to sentences like ``I think I have alcohol
  poisoning''  with the list of nearest liquor stores.}} they clearly
do not pass the test. 
One often forgets that, in addition to reasoning and access to
knowledge representation,  passing the Turing test presupposes
automated natural language analysis and generation which, 
despite significant progress in the field, 
has not yet been fully achieved. 
These natural language processing components of the Turing test are of independent interest and used in computer programs for question answering and translation (however, since both of these tasks are generally assumed to be AI-complete it is unlikely that a full solution for these problems would be simpler than a solution for the Turing test itself).

If we define the semantics of a (sequence of) sentence(s) $\sigma$ as the mapping to a representation 
$\phi(\sigma)$ that can be used by a machine for natural language processing tasks, two very different ideas of semantics come to mind. 

\begin{enumerate}
\item One notion of semantics describes what the sentence(s) speaks about. 
The dominant model for this type of semantics represents meaning using
word vectors (only involving referential/full words nouns, adjectives,
verbs, adverbs, \ldots  and not grammatical words) which represent what $\sigma$ speaks about. This is clearly computable. One must fix a thesaurus of $n$ words that acts as a vector basis. 
Usually words not in the thesaurus or basis are expanded into their definition 
with words in the thesaurus. By counting occurrences of words from the thesaurus in the text (substituting  words not in the thesaurus with their definition) 
and turning this into a $n$-dimensional vector reduced to be of euclidian norm $1$, we obtain word meanings in the form of $n$-dimensional vectors. 
This notion of semantics provides a useful measure of semantic
similarity between words and texts; typical applications include
exploring Big Data and finding relevant pages on the internet. This kind
of semantics  models what a word (or a text) speaks about. 

\item The other notion of semantics, the one this paper is about,  is of a
logical nature. It models what is asserted, refuted, \ldots  assumed
by the sentences. According to this view, computational semantics is
the mapping of sentence(s) to logical
formula(s). This is usually done compositionally, according to Frege's
principle \emph{``the meaning of a compound expression is a function
  of the meaning of its components''} to which Montague added
\emph{``and of its syntactic structure''}.  This paper focuses on this
logical and compositional notion of semantics and its extension (by us and others) to
lexical semantics; these extensions allow us to conclude from a
sentence like ``I started a book'' that the speaker started
\emph{reading} (or, depending on the context, writing) a  book.
\end{enumerate}

\editout{
Recently this notion of semantics has been extended (by us and others) in order to encompass some aspects of lexical semantics, that connect together predicates in the logical language. This is  indeed needed to reject (resp.\ to understand)  sentences like 

\begin{exe} 
\ex 
The table barked. 
\ex 
The sergeant barked. 
\ex
I started a new book. 
\end{exe} 
}


We should comment that, in our view, semantics is a (computable) function from  sentence(s) to logical formulae, since this viewpoint is not so common in  linguistics.

\begin{itemize} 
\item Cognitive sciences also consider the language faculty as a computational device and insist on the computations involved in language analysis and production. 
Actually there are two different views of this cognitive and 
computational view: one view, promoted by authors such as \cite{Pin94}, claims 
that there is a specific cognitive function for language, a ``language 
module'' in the mind, while others, like \cite{Langacker2008}, think that our language faculty is just our general cognitive abilities applied to language.
\item 
In linguistics and above all in philosophy of language many people think that sentences cannot have any meaning 
without a context, such a context involving both linguistic and extra-linguistic information. Thus, according to this 
view, the input of our algorithm should include context. Our answer is firstly that linguistic context is partly taken 
into account since we are able to produce, in addition to formulae, discourse structures. Regarding the part of context 
that we cannot take into account, be it linguistic or not, our answer is that it is not part of semantics, but rather an 
aspect of pragmatics. And, as argued by \cite{corblinsem}, if someone is given a few sentences on a sheet of 
paper without any further information, he starts imagining situations, 
may infer other statements from what he reads, \ldots, and such thoughts are the semantics of the sentence. 
\item 
The linguistic tradition initiated by \cite{montague} lacks some coherence regarding computability. On the one hand, Montague gives an algorithm for parsing sentences and for computing their meaning as a logical formula. 
On the other hand, he asserts that the meaning of a sentence is the interpretation of the formula in possible worlds, but these models are clearly uncomputable! Furthermore, according to him, each intermediate step, including the intensional/modal formulae should be forgotten, and the semantics is defined as the set of possible worlds in which the semantic formula is true: this cannot even be finitely described, except by these intermediate formulas; a fortiori it cannot be computed. Our view is different, for at least three reasons, from the weakest to the strongest: 
\begin{itemize} 
\item Models for higher order logic, as in Montague, are not as simple as is sometimes assumed, and they do not quite match the formulas: completeness fails. This means that a model and even all models at once contains less information than the formula itself. 
\item We do not want to be committed to any particular interpretation. Indeed, there are alternative relevant interpretations of formulas, as the 
following non exhaustive list shows: 
dialogical interpretations (that are the sets of proofs and/or refutations), 
game theoretic semantics and ludics (related to the former style of interpretation), 
set of consequences of the formula, 
structures inhabited by their normal proofs as in intuitionistic logic,... 
\item Interpreting the formula(s) is no longer related to linguistics, although some interpretations might useful for some applications. Indeed, once you have a a formula, interpreting it in your favourite way is a purely logical question. Deciding whether it is true or not in a model, computing all its proofs or all its refutations, defining game strategies, computing its consequences or the corresponding structure has nothing to do with the particular natural language statement you started with. 
\end{itemize} 
\end{itemize}

\section{Computational semantics \`a la Montague}

We shall first present the general algorithm that maps sentences to
logical formulae, returning to lexical semantics in
Section~\ref{sec:coercions}.  The first step is to compute a syntactic analysis that is rich and
detailed enough to enable the computation of the semantics (in the
form of logical formulae). The second step is to incorporate the lexical lambda terms 
and to reduce the obtained lambda term --- this step 
possibly includes the choice of some lambda terms from the lexicon that fix the type mismatches.

\subsection{Categorial syntax} 

In order to express the process that maps a sentence to its semantic
interpretation(s) in the form of logical formulae, we shall start with a
categorial grammar. This is not strictly necessary: \cite{montague} used a
 context free grammar (augmented with a mechanism for quantifier scope), but if one reads between
the lines, at some points he converts the phrase structure into
a categorial derivation, so we shall, following \cite{mr12lcg}, directly use a categorial
analysis. Although richer variants of categorial grammars are possible, and used in practice,
we give here an example with Lambek grammars, and briefly comment on variants later.



\newcommand\lex{\emph{lex}}
\newcommand\seq\vdash 

Categories are freely generated from  a set of base categories, for example $np$ (noun phrase), $n$ (common noun),
$S$ (sentence), by two binary operators: $\lto$ and $\lfrom$: 
$A\lto B$ and $B\lfrom A$ are categories whenever $A$ and $B$ are categories. A category $A\lto B$ intuitively looks for a category $A$ to its left in order to form a $B$. Similarly, a category $B\lfrom A$ combines with an $A$ to its right to form a $B$. The full natural deduction rules
are shown in Figure~\ref{fig:nd}.

\begin{figure}
\begin{align*}
\begin{prooftree} 
\Gamma\seq A\quad \Delta \seq A\lto B 
\justifies 
\Gamma, \Delta \seq  B 
\using \lto_e 
\end{prooftree} 
&&
\begin{prooftree} 
A, \Gamma\seq  B 
\justifies 
\Gamma\seq  A\lto B 
\using \lto_i 
\end{prooftree} 
\\
\begin{prooftree} 
\Delta  \seq B\lfrom A \quad \Gamma\seq A 
\justifies 
\Gamma, \Delta \seq  B 
\using \lto_e 
\end{prooftree} 
&&
\begin{prooftree} 
\Gamma, A\seq  B 
\justifies 
\Gamma\seq  B\lfrom A 
\using \lto_i 
\end{prooftree} 
\end{align*}
\caption{Natural deduction proof rules for the Lambek calculus}
\label{fig:nd}
\end{figure}

A lexicon provides, for each word $w$ of the language, a finite set of
categories $\lex(w)$. We say a sequence of words $w_1,\ldots, w_n$ is of type $C$ whenever
$\forall i \exists c_i \in \lex(w_i)\; c_1,\ldots,c_n\seq C$. Figure~\ref{fig:example} shows an example lexicon (top) and a derivation of
a sentence (bottom).
%
%
%
%
%
%
%
%
%
%
%
%
%
%
%
%
%

\newcommand\entree[2]
{\begin{array}[b]{c}
#1\\
{#2} 
\end{array} } 
\newcommand\obj{\[
\entree{\textit{a}}{((S\lfrom np)\lto S)\lfrom n}
\ 
\entree{\!\!\!\!\!\textit{cartoon}\!\!\!\!\!}{n} 
\justifies 
(S\lfrom np)\lto S
\using \lfrom_e
\] 
}
\newcommand\subj{
\[
\entree{\textit{every}}{(S\lfrom(np\lto S))\lfrom n}
\ \  
\entree{\textit{kid}}{n}
\justifies 
(S\lfrom(np\lto S))
\using \lfrom_e
\] 
} 
\newcommand\vvarobj[1]{
\[ 
\entree{\textit{watched}}{(np\lto S)\lfrom np }
\quad 
[np]^{#1}
\justifies 
(np\lto S) 
\using \lfrom_e
\] 
} 
\begin{figure}
\[
\begin{array}{r@{\ \ }l}
\textit{Word} & \textit{Syntactic Type} \\
\textrm{kid}  & n \\
\textrm{cartoon}  & n \\
\textrm{watched}  & (np\lto S)\lfrom np \\
\textrm{every} & (S\lfrom (np\lto S))\lfrom n \\
\textrm{a} & ((S\lfrom np)\lto S)\lfrom n \\
\end{array}
\]
\medskip
\[
\begin{prooftree} 
\[
\[
\subj
\vvarobj1
\justifies 
S
\using \lto_e
\]
\justifies 
S\lfrom np
\using \lfrom_i(1)
\]
\!\!\!\!\!\!\!\!\!\!
\obj
\justifies 
S
\using \lto_e
\end{prooftree}
\]
\caption{Lexicon and example derivation}
\label{fig:example}
\end{figure}

\subsection{From syntactic derivation to typed linear lambda terms} 

Categorial derivations, being a proper subset of derivations in
 multiplicative intuitionistic linear logic, correspond to (simply
 typed) linear lambda terms. This makes the connection to Montague grammar particularly transparent.

Denoting by $e$ the set of entities (or individuals) and by $t$ the
type for propositions (these can be either true or false, hence the name $t$)
one has the following mapping from syntactic categories to
semantic/logical types.
\[
\begin{array}{|rclp{0.6\textwidth}|}\hline  	
(\makebox{Syntactic type})^* &=& \multicolumn{2}{l|}{\makebox{Semantic 
type}}\\ \hline 
	S^{*} &=& t & a sentence is a proposition \\ 
	np^{*} &=& e & a noun phrase is an entity \\
	n^{*} &=& e\imp t & a noun is a subset of the set of entities (maps entities to  propositions) \\ 
	(A\lto B)^*=(B\lfrom A)^* & = & A^* \fl B^* & extends easily  to all syntactic categories \\  \hline 
\end{array} 
\]
Using this translation of categories into types which forgets the non commutativity, 
the Lambek calculus proof of Figure~\ref{fig:example} is translated to the linear intuitionistic proof shown in
Figure~\ref{fig:exll}; we have kept the order
of the premisses unchanged to highlight the similarity with the
previous proof.  Such a proof can be viewed as a simply typed lambda term with the two base types $e$ and $t$.
\[ (a^{(e\imp t)\imp ((e\imp t) \imp t)}\ cartoon^{e\imp t})(\lambda
y^\eee (every^{(e\imp t)\imp ((e\imp t) \imp t)}\ kid^{e\imp
  t})(watched^{e\imp e\imp t}\ y))\]

\begin{figure}
\newcommand\objs{\[ 
\entree{a}{(\eee\flsshort\ttt)\flsshort(\eee\flsshort\ttt)\flsshort\ttt}
\  
\entree{cartoon}{(\eee\flsshort\ttt)}
\justifies 
(\eee\flsshort\ttt)\flsshort\ttt
\using \flsshort_e
\] 
}

\newcommand\subjs{\[ 
\entree{every}{(\eee\flsshort\ttt)\flsshort(\eee\flsshort\ttt)\flsshort\ttt}
\
\entree{kid}{(\eee\flsshort\ttt)} 
\justifies 
(\eee\flsshort\ttt)\flsshort\ttt
\using \flsshort_e
\] 
}

\newcommand\vvarobjs[1]{
\[ 
\entree{watched}{\eee\flsshort\eee\flsshort\ttt}
\ \  
\entree{y}{[\eee]^{#1}}
\justifies 
\eee\flsshort\ttt
\using \fls_e
\] 
}

\begin{center} 
\scalebox{.8}{
\begin{prooftree} 
\[
\[ 
\subjs
\vvarobjs1
\justifies 
\ttt
\using \flsshort_e
\] 
\justifies 
\eee\flsshort\ttt
\using \flsshort_{i(1)}
\]
\!\!\!\!\!\!\!\!
\objs
\justifies 
\ttt
\using \flsshort_e
\end{prooftree}}
\end{center}
\vspace{-.5\baselineskip}
\caption{The multiplicative linear logic proof corresponding to
  Figure~\ref{fig:example}}
\label{fig:exll} 
\end{figure}

\noindent


As observed by \cite{church40}, the simply typed lambda calculus with two types $e$ and $t$ 
is enough to express  higher order logic, provided one introduces 
\emph{constants} for the logical connectives and quantifiers, that is
a constants ``$\exists$'' and ``$\qqst$''  of type $(e \imp t) \imp t$, and
constants ``$\et$'', ``$\ou$'' et ``$\implique$'' of type $t \imp (t \imp t) $.

%
%

In addition to the syntactic lexicon, there is a semantic lexicon that maps
any word to a simply typed lambda term with atomic types $e$ and $t$
and whose type is the translation of its syntactic formula. Figure~\ref{fig:semlex} presents such a lexicon for our current example. For
example, the word ``every'' is assigned formula  $(S\lfrom (np \lto
S))\lfrom n$. According to the translation function above, we know the
corresponding semantic term must be of type $(e\imp t)\imp ((e\imp t)
\imp t)$, as it is in Figure~\ref{fig:exll}.
The term we assign in in the semantic lexicon is the following (both
the type and the term are standard in a Montagovian setting).
\[
\l P\type{e\imp t}\  \l Q\type{e\imp t}\  
(\qqst\type{(e\imp t)\imp t}\  (\l x\type{e}  (\Rightarrow\type{t\imp (t\imp t)} (P\ x) (Q\ x))))
\]
Unlike the lambda terms computed for proof, the lexical entries in the semantic lexicon need not be linear: the lexical entry above is not a linear lambda term since the single abstraction binds two occurrences of $x$.

Similarly, the syntactic type of ``a'', the formula $((S\lfrom np) \lto
S)\lfrom n$ has corresponding semantic type $(e\imp t)\imp ((e\imp t)
\imp t)$ (though syntactically different, a subject and an object
generalized quantifier have the same \emph{semantic} type), and the
following lexical meaning recipe.
\[
\l P\type{e\imp t}\  \l Q\type{e\imp t}\  
(\existe\type{(e\imp t)\imp t}\  (\l x\type{e}  (\et\type{t\imp (t\imp t)} (P\ x) (Q\ x))))
\]

Finally, ``kid'', ``cartoon'' and ``watched'' are assigned the
constants $\texttt{kid}\type{e\imp t}$, $\texttt{cartoon}\type{e\imp
  t}$ and $\texttt{watched}\type{e \imp (e \imp t)}$ respectively.

\begin{figure}
\begin{tabular}{ll} \hline 
\textbf{word} & \textbf{\itshape syntactic type $u$}\\ 
&  \textbf{\itshape semantic type $u^*$}\\ 
& \textbf{\itshape  semantics: $\l$-term of type $u^*$}\\ \hline 
every & $(S\lfrom (np \lto S))\lfrom n$ (subject) \\ 
& $(e\imp t)\imp ((e\imp t) \imp t)$\\ 
& $\l P\type{e\imp t}\  \l Q\type{e\imp t}\  
(\qqst\type{(e\imp t)\imp t}\  (\l x\type{e}  (\Rightarrow\type{t\imp (t\imp t)} (P\ x) (Q\ x))))$ \\  \hline 
a  & $((S\lfrom np) \lto S)\lfrom n$  (object)\\ 
& $(e\imp t)\imp ((e\imp t) \imp t)$\\ 
& $\l P\type{e\imp t}\  \l Q\type{e\imp t}\  
(\existe\type{(e\imp t)\imp t}\  (\l x\type{e}  (\et\type{t\imp (t\imp t)} (P\ x) (Q\ x))))$ \\  \hline 
kid  & $n$ \\
& $e\imp t$\\ 
& $\l x\type{e} (\texttt{kid}\type{e\imp t}\  x)$\\  \hline 
cartoon  & $n$ \\
& $e\imp t$\\ 
& $\l x\type{e} (\texttt{cartoon}\type{e\imp t}\  x)$\\  \hline 
watched & $(np \lto S)\lfrom np$ \\ 
& $e\imp (e \imp t)$\\ 
& $\l y\type{e}\  \l x\type{e}\  ((\texttt{watched}\type{e \imp (e \imp t)}\  x) \ y)$ \\  \hline 
\end{tabular} 
\caption{Semantic lexicon for our example grammar}
\label{fig:semlex}
\end{figure}

Because the types of these lambda terms are the same as those of the
words in the initial lambda term, we can take the linear lambda term
associated with the sentence and substitute, for each word its
corresponding lexical meaning, transforming the
derivational semantics, in our case the following\footnote{There are
\emph{exactly} two (non-equivalent) proofs of this sentence.
The second proof using the same premisses corresponds to the
second, more prominent reading of the sentence whose lambda term is: 
$(every\ kid)  (\l x^\eee. (a\ cartoon)(\lambda y^\eee ((watched\ y)\ x))$}
\[
(a^{(e\imp t)\imp ((e\imp t) \imp t)}\ cartoon^{e\imp t})(\lambda
y^\eee (every^{(e\imp t)\imp ((e\imp t) \imp t)}\ kid^{e\imp
  t})(watched^{e\imp e\imp t}\ y))
\]

\noindent into an (unreduced) representation of the meaning of the sentence.
\begin{align*}
((\l P\type{e\imp t}\  \l Q\type{e\imp t} 
(\existe\type{(e\imp t)\imp t}\  (\l x\type{e}  (\et\type{t\imp (t\imp t)} (P\ x) (Q\ x)))))
\,\texttt{cartoon}\type{e\imp t})
\\
((\lambda y^\eee (
((\l P\type{e\imp t}\  \l Q\type{e\imp t}\  
(\qqst\type{(e\imp t)\imp t}\  (\l x\type{e}  (\Rightarrow\type{t\imp (t\imp t)} (P\ x) (Q\ x)))))
\,\texttt{kid}\type{e\imp t}\  x))) (\texttt{watched}\type{e\imp e\imp t} y))
\end{align*}

The above term reduces to 
\[
(\existe\type{(e\imp t)\imp t}\ \l x^e (\et\type{t\imp (t\imp t)}
(\texttt{cartoon}\  x) (\qqst\type{(e\imp t)\imp t}\ (\l z^e (\Rightarrow\type{t\imp (t\imp t)} (\texttt{kid}\ z)
((\texttt{watched}\  x)\, z))))))
\]

\noindent that is\footnote{We use the standard convention to translate
  a term $((p\, y)\, x)$ into a predicate $p(x,y)$.}:\quad 
$\exists x. \texttt{cartoon}(x) \wedge \forall z. \texttt{kid}(z)
\Rightarrow \texttt{watched}(z,x)
$ 
%
%
\editout{
\subsection {Semantics: direct homomorphism}

\[
\begin{array}{ccc}
\infer[/E]{A:(f\ x)^T}{
A/B:f^{U\rightarrow T} & B:x^U} &&
\infer[/I]{A/B:\lambda x^U t}{\infer*{A:t^T}{\ldots [B:x^U]}} \\[5mm]
\infer[\backslash E]{A:(f\ x)^T}{
B:x^U  & B\backslash A:f^{U\rightarrow T}} &&
\infer[\backslash I]{B\backslash A:\lambda x^U t}{\infer*{A:t^T}{[B:x^U]\ldots}} \\[5mm]
\end{array}
\]
}

\begin{figure}
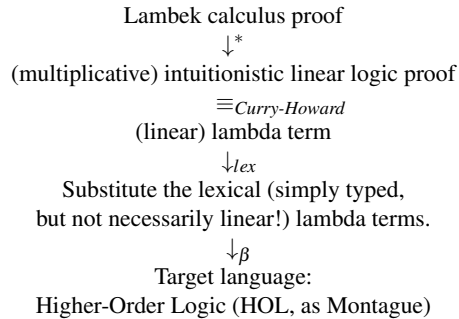

\begin{center} 
\begin{tabular}{c}
\\ Lambek calculus proof \\
\smash{$\downarrow^*$} \\
(multiplicative) intuitionistic linear logic proof\\
$\qquad\qquad \equiv_{\textit{Curry-Howard}}$ \\
(linear) lambda term
\\
\ \ \smash{$\downarrow_{lex}$} \\ 
Substitute the lexical (simply typed,\\ but not necessarily linear!) lambda terms.\\
\ \ \smash{$\downarrow_{\beta}$} \\
Target language:\\ Higher-Order Logic (HOL, as Montague)\\
\end{tabular}
\end{center}
\caption{The standard categorial grammar method for computing meaning} 
\label{fig:old}
\end{figure}

The full algorithm to compute the semantics of a sentence as a
logical formula is shown in Figure~\ref{fig:old}.

\section{Adding sorts, coercions, and uniform operations}
\label{sec:coercions}

Montague (as Frege) only used a single type for entities:  $e$.
But it is much better to have many sorts in order to block 
the interpretation of some sentences:

\begin{exe} 
\ex * The table barked.
\ex The dog barked. 
\ex ?The sergeant barked. 
\end{exe} 

As dictionaries say \ma{barked}  can be  said from animals, usually dogs. 
The first one is correctly rejected: one gets 
$\texttt{bark}^{\textit{dog}\imp t}(\texttt{the}\ \texttt{table})^{\textit{artifact}}$ and $\textit{dog}\neq \textit{artifact}$. 

However we need to enable the last example $\texttt{bark}^{\textit{dog}\imp t}
(\texttt{the}\ \texttt{sergeant})^{\textit{human}}$ 
and in this case we use coercions \citep{bmr10tt,Retore2014types2013}:
the lexical entry for the verb \ma{barked} which only applies to the sort of \ma{dogs} provides a coercion $c: \textit{human}\imp \textit{dog}$ from \ma{human}  to \ma{dog}. 
The revised lexicon provides each word with the lambda term that we saw earlier 
(typed using some of the several sorts / base type) and some optional lambda terms that can be used if needed to solve type mismatches. 

Such coercions are needed to understand sentences like:
\begin{exe}
\ex This book is heavy.
\ex This book is interesting. 
\ex This book is heavy and interesting. 
\ex Washington borders the Potomac. 
\ex Washington attacked Iraq. 
\ex * Washington borders the Potomac and attacked Iraq. 
\end{exe}

\begin{figure}
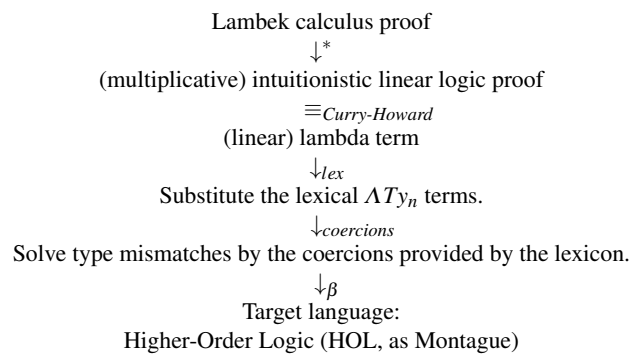

\begin{center}
\begin{tabular}{c}
\\ Lambek calculus proof \\
\smash{$\downarrow^*$} \\
(multiplicative) intuitionistic linear logic proof\\
$\qquad\qquad \equiv_{\textit{Curry-Howard}}$ \\
(linear) lambda term
\\
\ \ \smash{$\downarrow_{lex}$} \\ 
Substitute the lexical $\Lambda Ty_n$ terms.\\
\ \ \ \  \ \ \ \ \ \, \smash{$\downarrow_{coercions}$} \\ 
Solve type mismatches by the coercions provided by the lexicon. \\
\ \ \smash{$\downarrow_{\beta}$} \\
Target language:\\ Higher-Order Logic (HOL, as Montague)\\
\end{tabular}
\end{center}
\caption{Computing meaning in a framework with coercion}
\label{fig:new}
\end{figure}

The first two sentences will respectively use a coercions from book to
physical object and a coercion from books to information. Any time an
object has several related meanings, one can consider the conjunction
of properties referring to those particular aspects. For these operations (and others acting uniformly on types) we exploit polymorphically typed lambda terms (system F). 
When the related meanings of a word  are incompatible (this is usually
the case) the corresponding coercions are declared to be incompatible in the lexicon (one is declared as rigid). This extended process is described in Figure~\ref{fig:new}. Some remarks on our use of system F: 
\begin{itemize} 
\item We use it  for the syntax of semantics (a.k.a. metalogic, glue logic)
\item The formulae of semantics are the usual ones (many sorted as in $Ty_n$)
\item We have a single constant for operations that act uniformly on types, like quantifiers or conjunction over predicates that apply to different facets of a given word. 
\end{itemize} 
%
%
%
%
%
%
%
%
%
%
%
%
%
%
%
%
%
%
\editout{
\subsection {System F} 
%
%
%
%
%
The types are defined as follows: 
\begin{itemize} 
\item 
Constants $e$ (or $e_1,\ldots,e_n$ in a multisorted system) and $t$, as well as any type variable $\alpha$ in $P$, are types. 
\item 
Whenever $T$ is a type and $\alpha$ a type variable which may but need not occur in $T$,  $\Pi. \alpha.\ T$ is a type. 
\item 
Whenever $T_1$ and $T_2$ are types, $T_1\fl T_2$ is also a type.
\end{itemize}

The typed terms are defined as follows: 
\begin{itemize} 
\item A variable  of type $T$ i.e. $x:T$ or  $x^T$  is a \emph{term}.\\  Countably many variables of each type.
\item 
$(f\ \tau)$ is a term of type $U$ whenever $\tau:T$ and  $f:T\fl U$. 
\item 
$\lambda x^T\!\!.\ \tau$ is a term of type $T\fl U$ whenever $x:T$, 
and $\tau:U$.  
\item $\tau \{U\}$ is a term of type $T[U/\alpha]$
whenever $\tau:\Pi \alpha.\ T$, and $U$ is a type. 
\item $\Lambda \alpha. \tau$ is a term of type $\Pi \alpha. T$
whenever $\alpha$ is a type variable, and  $\tau:T$ without any free occurrence of the type variable $\alpha$.  
\end{itemize}

Finally, reduction is defined as follows: 
\begin{itemize} 
\item $(\Lambda \alpha. \tau) \{U\}$  reduces to $\tau[U/\alpha]$ (remember that $\alpha$ and $U$ are types). 
\item $(\lambda x. \tau) u$ reduces to $\tau[u/x]$ (usual reduction). 
\end{itemize} 
}
%
%

%
%
%


\section{Complexity of the syntax} 
\label{sec:syntax}





As we remarked before, when computing the formal semantics of a sentence in the Montague tradition, we (at least implicitly) construct a categorial grammar
proof. Therefore, we need to study the complexity of parsing/theorem proving in categorial grammar  first. 
The complexity generally studied in this context is the complexity of deciding about the existence of a proof (a parse) for a logical statement (a natural language sentence) as a function of the number of words in this sentence\footnote{For many algorithms, the complexity is a function of the number of atomic subformulas of the formulas in the sentence. Empirically estimation shows the number of atomic formulas is a bit over twice the number of words in a sentence.}.

Perhaps surprisingly, the simple product-free version of the Lambek
calculus we have used for our examples is already NP-complete
\citep{savateev}. However, 
there is a notion of
order, which measures the level of ``nesting'' of the
implications as defined below. 
\begin{align*}
\textit{order}(p)  &  = 0 \\
\textit{order}(A/B) = \textit{order}(B\backslash A) 
& = \textrm{max}(\textit{order}(A),(\textit{order}(B)+1))
\end{align*}

As an example, the order of formula $(np\backslash S)/np$ is 1, whereas the order of formula $S/(np\backslash S)$ is 2.
For the Lambek calculus, the maximum order of the formulas in a grammar is
 a good indication of its complexity. Grammars used for
linguistic purposes generally have formulas of order 3 or, at most, 4. We know that once we bound the 
order of formulas in the lexicon of our grammars to be less than a fixed $n$,  parsing becomes polynomial for any choice of $n$  \citep{pentus10poly}\footnote{For the algorithm of
  \cite{pentus10poly}, the order appears as an exponent in the
  worst-case complexity: for a grammar of order $n$ there is a multiplicative factor of
  $2^{5(n+1)}$. So though polynomial, this algorithm is not necessarily efficient.}.

The NP-completeness proof of \cite{savateev} uses a reduction from SAT,
where a SAT problem with $c$ clauses and $v$ variables produces a
Lambek grammar of order $3+4c$, with $(2c+1)(3v +1)$ atomic
formulas\editout{\footnote{Though reducing the number of atomic formulas
  increases the nondeterminism in proof search, the NP completeness
  proof uses these distinct atomic formulas to ensure all available choices in
  proof search correspond to an assignment of true or false to a
  specific variable in the original problem.}}. 

The notion of order therefore provides a neat
indicator of the complexity: the NP-completeness proof requires
formulas of order 7 and greater, whereas the formulas used for
linguistic modelling are of order 4 or less.

Even though the Lambek calculus is a nice and simple system, we know that the Lambek calculus generates only
context-free languages \citep{pentus}, and there is good evidence
that at least some constructions in natural language require a
slightly larger class of languages \citep{shieber}. One influential
proposal for such a larger class of languages are the mildly
context-sensitive languages \citep{tagcs}, characterised as follows. 
\begin{itemize}
\item contains the context-free languages,
\item limited cross-serial dependencies (i.e includes $a^nb^nc^n$ but maybe not $a^nb^nc^nd^ne^n$)
\item semilinearity (a language is semilinear iff there exists a regular
  language to which it is equivalent up to permutation)
\item polynomial fixed recognition\footnote{The last two items are
    sometimes stated as the weaker condition ``constant growth'' instead of
    semilinearity and the stronger condition of polynomial parsing
    instead of polynomial fixed recognition. Since all other
    properties are properties of formal languages, we prefer the formal language theoretic notion of polynomial fixed recognition.}
\end{itemize}

There are various extensions of the Lambek calculus which generate mildly context-sensitive languages while keeping the syntax-semantics interface essentially the same as for the Lambek calculus. Currently,  little is known about upper bounds of the classes of formal 
languages generated by these extensions of the Lambek calculus. Though 
 \cite{diss} shows that multimodal categorial grammars generate exactly the 
context-sensitive languages, \cite{Bus97} underlines the difficulty of adapting the result of 
\cite{pentus} to extensions of the Lambek calculus\footnote{We can side-step the need for a Pentus-like proof by looking only at fragments of order 1, but these fragments are insufficient even for handling quantifier scope.}.

Besides problems from the point of view of formal language theory, it should be noted that the goal we set out at the start of this paper
was not just to generate the right string language but rather to
generate the right string-meaning pairs. This poses additional
problems. For example, a sentence with $n$ quantified noun phrases has
up to $n!$ readings.
Although the standard notion of complexity for categorial grammars is the complexity deciding whether or not a proof exists, formal semanticists, at least since
\cite{montague}, want their formalisms to generate all and only the correct readings for a sentence: we are not only interested in whether or not
a proof exists but, since different natural deduction proofs correspond to different readings, also in what the different proofs of a sentence are\footnote{Of course, when our goal is to generate (subsets of) $n!$
  different proofs rather than a single proof (if one exists), then we are no longer in NP, though it is unknown whether an algorithm exists which produces a sort of shared representation for all such subsets such
  that 1) the algorithm outputs ``no'' when the sentence is ungrammatical 2) the algorithm has a fairly trivial algorithm (say of a low-degree polynomial at worst) for
  recovering all readings from the shared representation 3) the shared structure is polynomial in the size of the input.}.

When we look at the example below

\begin{exe}
\ex\label{ex:samples} Every representative of a company saw most samples. 
\end{exe}

\noindent it has five possible readings
(instead of $3! = 6$), since the reading 
\[
\forall x.
\texttt{representative\_of}(x,y) \Rightarrow
\texttt{most}(z,\texttt{sample}(z)) \Rightarrow \exists
y. \texttt{company}(y)\ \wedge \texttt{see}(x,z)
\] 

\noindent has an unbound occurrence of $y$ (the leftmost occurrence). The Lambek calculus analysis has
trouble with all readings where ``a company'' has wide
scope over at least one of the two other quantifiers. We can, of course, remedy this by adding new, more complex types to the
  quantifier ``a'', but this would increase the order of the formulas
  and there is, in principle, no bound on the number of constructions
  where a medial quantifier has wide scope over a sentence. A simple counting argument shows that Lambek calculus grammars cannot generate the  $n!$ readings required
for quantifier scope of an $n$-quantifier sentence: the number of readings for a Lambek calculus proof is proportional to the Catalan numbers and this number is in $o(n!)$\footnote{We
  need to be careful here: the number of readings
  for a sentence with $n$ quantifiers is $\Theta(n!)$, whereas the maximum number of Lambek calculus proofs is $O(c_0^{c_2n} C_{c_1c_2n})$, for constants
  $c_0$, $c_1$, $c_2$ which depend on the grammar ($c_0$ is the maximum number of formulas for a single word, $c_1$ is the maximum number of (negative) atomic subformulas for a
  single formula and $c_2$ represent the minimum number of words needed to add a generalized quantifier to a sentence, i.e.\ $c_2n$ is the number of words required to produce an
  $n$-quantifier sentence) and $O(c_0^{c_2n} C_{c_1c_2n})$ is in $o(n!)$.}; in other words,
given a Lambek calculus grammar, the number of readings of a sentence with $n$ quantifiers grows
much faster than the number of Lambek calculus proofs for this sentence, hence the grammar fails to generate many of the required readings.

Since the eighties, many variants and extensions of the Lambek
calculus have been proposed, each with the goal of overcoming the limitations of the Lambek calculus. 
Extensions/variations of the Lambek calculus --- which include multimodal
categorial grammars \citep{M95}, the Displacement calculus \citep{mvf11displacement} and first-order linear
logic \citep{mill1} --- solve both the problems of formal language theory and the
problems of the syntax-semantics interface. For example, there are
several ways of implementing quantifiers 
yielding exactly the five desired readings for sentence~\ref{ex:samples} 
without appealing to extra-grammatical
mechanisms. \cite{carpenter1994deductive} gives many examples of
the advantages of this logical approach to scope, notably its interaction with other
semantic phenomena like negation and coordination. 

Though these modern calculi solve the problems with the Lambek
calculus, they do so without excessively increasing the computational
complexity of the formalism: multimodal categorial grammars are PSPACE
complete \citep{diss}, whereas most other extensions are NP-complete, like the
Lambek calculus. 

Even the most basic categorial grammar
account of quantifier scope requires formulas of order 2, while, in contrast to the Lambek calculus, the only known polynomial
fragments of these logics are of order 1. 
Hence the known polynomial fragments have very limited
appeal for semantics. 

Is the NP-completeness of our logics in conflict with the condition of
polynomial fixed recognition required of mildly context-sensitive
formalisms? Not necessarily, since our goals are different: we are not only interested in the string language generated by our formalism but also in the string-meaning
mappings. 
Though authors have worked on using 
 mildly context-sensitive formalisms
for semantics, they generally use one of the two following strategies
for quantifier scope: 1) an external mechanism for computing quantifier scope (e.g.\
  Cooper storage, \citep{cooper75}), or 2) an underspecification mechanism for representing quantifier scope \citep{fl10under}.

For case 1 \citep{cooper75}, a single syntactic structure is converted into up to
$n!$ semantic readings, whereas for case 2, though we represent all
possible readings in a single structure, even
deciding whether the given sentence has a semantic reading \emph{at all}
becomes NP-complete \citep{fl10under}, hence we simply shift the NP-completeness from
the syntax to the syntax-semantics interface\footnote{In addition, \citet{ebert05phd} argues that underspecification languages are not expressive enough to capture all possible readings of a sentence in a single structure. So underspecification does not solve the combinatorial problem but, at best, reduces it.}. Our current understanding therefore indicates that
NP-complete is the best we can do when we want to generate the
semantics for a sentence. We do not believe this to be a bad
thing, since pragmatic and processing constraints rule out many of the
complex readings and enumerating all readings of sentences like
sentence~\ref{ex:samples} above (and more complicated examples) is a
difficult task. There is a trade-off between the work done in the
syntax and in the syntax-semantics interface, where the categorial
grammar account incorporates more than the traditional mildly
context-sensitive formalisms. It is rather easy to set up a categorial
grammar parser in such a way that it produces underspecified
representations in time proportional to $n^2$ \citep{moot07filter}. However, given that such an underspecified
representation need not have any associated semantics, such a system
would not actually qualify as a parser. We believe,
following \cite{carpenter1994deductive} and \cite{j02disorg}, that
giving an integrated account of the various aspects of the
syntax-semantics interface is the most promising path.

Our grammatical formalisms are not merely theoretical tools, but also
form the basis of several implementations \citep{mv15cctlg,moot15l1}, with a rather extensive
coverage of various semantic phenomena and their interactions,
including quantification, gapping, ellipsis, coordination, comparative
subdeletion, etc.

\section{Complexity of the semantics}
\label{sec:semantics}

The complexity of the syntax discussed in the previous section only
considered the complexity of 
computing unreduced lambda terms as the meaning of a sentence. Even
in the standard, simply typed Montagovian framework, normalizing
lambda terms is known to be of non-elementary complexity \citep{sch82complex}, essentially
due to the possibility of recursive copying. In spite of this
forbidding worst-time complexity, normalization does not seem to be a
bottleneck in the computation of meaning for practical applications \citep{bos04sem,moot10grail}. 

Is there a deeper reason for this? We believe that natural language
semantics uses a restricted fragment of the lambda calculus,
soft lambda calculus. This calculus restricts recursive
copying and has been shown to characterize the complexity class P
exactly \citep{lafont04soft,baillot04soft}. Hence, this would explain why even naive implementations of
normalization perform well in practice.

The question of whether soft linear logic suffices for our semantic parser may appear hard to answer, however, it an obvious (although tedious) result. 
To show that all the semantic lambda terms can be typed in soft linear logic, we only need to verify that every lambda in the lexicon is soft. There is a finite number of words, with only a finite number of lambda terms per word. Furthermore, words from open classes (nouns, verbs, adjectifs, manner adverbs,... in which speakers may introduce new words... about 200.000 inflected word forms) are the most numerous and all have soft and often even linear lambda terms. Thus only closed class words (grammatical words such as pronouns, conjunctions, auxiliary verbs,... and some complex adverbs, such as ``too'') may potentially need a non-soft semantic lambda term: there are less than 500 such words, so it is just a matter of patience to prove they all have soft lambda terms. Of course, finding deep reasons (cognitive, linguistic) for semantic lambda terms to be soft in \emph{any} language would be much more difficult (and much more interesting!).

When adding coercions, as in Section~\ref{sec:coercions}, the
process becomes a bit more complicated.  However, the system of
\cite{lafont04soft} includes second-order quantifiers hence reduction
stays polynomial once coercions have been chosen. Their choice (as the choice of the syntactic category) 
increases complexity: 
when there is a type mismatch $g^{A\fl X} u^B$ one needs to chose one of the coercions of type $B\fl A$ provided by the entries of the words in the analysed phrase, with
the requirement that when a rigid coercion is used, all other coercions provided by the same word are blocked (hence rigid coercions, as opposed to flexible coercions decrease the number of choices for other type mismatches). 

%







Finally, having computed a set of formulas in higher-order logic corresponding
to the meaning of a sentence, though of independent interest for formal semanticists, is only a step towards using these meaning representations for concrete applications.
Typical applications such as question answering, automatic summarization, etc.\   require world knowledge and common sense reasoning but also a method for deciding about entailment: that is, given a set of sentences, can we conclude that
another sentence is true. This question is of course undecidable,
already in the first-order case. However, some recent research
shows that even higher-order logic formulas of the type produced by
our analysis can form the basis of effective reasoning mechanisms
\citep{cl14coq,mineshimahigher} and we leave it as an interesting open
question to what extent  such reasoning can be applied to natural language
processing tasks.

\editout{
\section{Other complexity measures} 


Is worst case complexity a good measure for natural language processing? 
Though we should be careful about making a
direct connection between human cognitive processing and our parsing algorithms, 
we should prefer a model in which 
sentences~\ref{cheesea} and~\ref{twoa} are easier to parse than sentences~\ref{cheeseb} and~\ref{twob}: 
\begin{exe}
\ex\label{cheesea} the cat that saw the rat that ate the cheese that stank 
\ex\label{cheeseb} the cheese that the rat that the cat saw ate stank
\ex\label{twoa} Frank was bothered that Ingrid was astonished that Jack was surprised that two plus two equals four 
\ex \label{twob} that that that two plus two equals four surprised Jack astonished Ingrid bothered Frank 
\end{exe}
}
\section{Conclusion} 

It is somewhat surprising that, in constrast to the well-developed theory of the algorithmic complexity of parsing, little is known about semantic analysis, 
even though computational semantics is an active field, as the recurring conferences with the same title as well as the number of natural language processing applications show. 
In this paper we simply presented remarks on the computability and on the complexity of this process. 
The good news is that semantics (at least defined as a set of logical formula) is \emph{computable}. 
This was known, but only implicitly: Montague gave a set of instruction to compute the formula (and to interpret it in a model), 
but he never showed that, when computing such logical formula(s): 
\begin{itemize} 
\item the process he defined stops with a normal lambda terms of type proposition ($\ttt$),
\item eta-long normal lambda terms with constants being either logical connectives or constants of a first (or higher order) logical language are in bijective correspondence with formulas of this logical language (this is more or less clear in the work of \citet{church40} on simple type theory).
\item the complexity of the whole process has a known complexity class, in particular the beta-reduction steps which was 
only discovered years after his death \citep{sch82complex}.
\end{itemize} 

A point that we did not discuss is that we considered \emph{worst case complexity} viewed as a function from the \emph{number of words in a sentence} a logical
formula. Both aspects of our point of view can be challenged: in practice, grammar size is at least as importance as sentence length and average case complexity may be
more appropriate than worst case complexity. 
Though the high worst case complexity shows that computing the semantics of a sentence is not 
always efficient, 
we nevertheless believe, confirmed by actual practice, that 
statistical models of a syntactic or semantic domain improve efficiency considerably, 
by providing extra information (as a useful though faillible ``oracle'') for many of the difficult choices. 
Indeed, 
human communication and understanding are very effective in general, 
but, from time to time, we misunderstand eachother or need to ask 
for clarifications. For computers, 
the situation is almost identical: most sentences are analysed 
quickly, while some require more time or even defeat the software. 
Even though it is quite difficult to obtain the 
actual probability distribution on sentence-meaning 
pairs, we can simply estimate such statistics empirically by randomly selecting manually annotated examples from a corpus. 
The other aspect, the sentence length, is, as opposed to what is commonly assumed in complexity theory, not a very safisfactory empirical measure of performance: indeed the average number of words per sentence is
around 10 in spoken language and around 25 in written language. Sentences with more than 100 words are very rare\footnote{To given an indication, the TLGbank contains more than
  14.000 French sentences and has a median of 26 words per sentence, 99\% of sentences having less than 80 words, with outliers at 190 and at 266 (the maximum sentence
  length in the corpus).}. Furthermore, lengthy sentences tend to have a simple structure, because otherwise they would quickly become incomprehensible (and hard to
produce as well). Experience with parsing shows that in many cases, the grammar size is at least as important as sentence length for the empirical complexity of parsing
algorithms \citep{joshi97parsing,sarkar00practical,rodr06parsers}. Grammar size, though only a constant factor in the complexity, tends to be a \emph{big} constant for realistic
grammars: grammars with between 10.000 and 20.000 rules are common.


We believe that the complexity of computing the semantics and of reasoning with the semantic representations are some of the most important reasons that 
the Turing test is presently out of reach. 

\bibliography{moot}
\bibliographystyle{agsm}



\end{document}